\documentclass[a4paper,11pt]{article}

\usepackage[margin=1in]{geometry} 
\usepackage[utf8]{inputenc}
\usepackage{authblk}
\usepackage{titlesec} 
\usepackage{setspace} 
\usepackage{graphicx}
\usepackage[style=numeric,sorting=none]{biblatex}
\usepackage{comment}
\usepackage{float}
\usepackage{amsmath}
\usepackage{amssymb}
\usepackage{makecell,color}
\usepackage{booktabs}

\definecolor{jcolor}{RGB}{041,122,000}
\definecolor{darkblue}{RGB}{000,000,150}
\definecolor{darkred}{RGB}{100,000,000}
\definecolor{purple}{RGB}{200,000,200}

\titleformat{\section}{\bfseries}{\thesection.}{1em}{}
\titleformat{\subsection}{\normalsize\bfseries}{\thesubsection.}{1em}{}

\title{\Large\textbf{A Weakly Supervised Transformer for Rare Disease Diagnosis and Subphenotyping from EHRs with Pulmonary Case Studies}}

\author[1]{Kimberly F. Greco$^{*}$}
\author[2]{Zongxin Yang$^{*}$}
\author[3]{Mengyan Li}
\author[4]{Han Tong}
\author[2]{Sara Morini Sweet}
\author[5,6,7]{Alon Geva}
\author[7,8]{Kenneth D. Mandl$^{\dag}$}
\author[9,10]{Benjamin A. Raby$^{\dag}$}
\author[1,2]{Tianxi Cai$^{\dag}$}

\affil[1]{\scriptsize Department of Biostatistics, Harvard T.H. Chan School of Public Health, Boston, USA}
\affil[2]{Department of Biomedical Informatics, Harvard Medical School, Boston, USA}
\affil[3]{Department of Mathematical Sciences, Bentley University, Waltham, USA}
\affil[4]{Department of Statistics, Columbia University, New York, USA}
\affil[5]{Department of Anesthesiology, Critical Care, and Pain Medicine, Boston Children's Hospital, Boston, USA}
\affil[6]{Department of Anesthesia, Harvard Medical School, Boston, USA}
\affil[7]{Computational Health Informatics Program, Boston Children's Hospital, Boston, USA}
\affil[8]{Department of Pediatrics, Harvard Medical School, Boston, USA}
\affil[9]{Division of Pulmonary Medicine, Boston Children's Hospital, Harvard Medical School, Boston, USA}
\affil[10]{Channing Division of Network Medicine, Brigham and Women's Hospital, Harvard Medical School, Boston, USA}

\date{}

\begin{document}

\maketitle

\footnotetext[1]{These authors contributed equally to this work.}
\footnotetext[2]{These authors jointly supervised this work.}

\section*{Abstract} 

Rare diseases affect an estimated 300-400 million people worldwide, yet individual conditions remain underdiagnosed and poorly characterized due to their low prevalence and limited clinician familiarity. Computational phenotyping offers a scalable approach to improving rare disease detection, but algorithm development is hindered by the scarcity of high-quality labeled data for training. Expert-labeled datasets from chart reviews and registries are clinically accurate but limited in scope and availability, whereas labels derived from electronic health records (EHRs) provide broader coverage but are often noisy or incomplete. To address these challenges, we propose WEST (\textbf{WE}akly \textbf{S}upervised \textbf{T}ransformer for rare disease phenotyping and subphenotyping from EHRs), a framework that combines routinely collected EHR data with a limited set of expert-validated cases and controls to enable large-scale phenotyping. At its core, WEST employs a weakly supervised transformer model trained on extensive probabilistic silver-standard labels -- derived from both structured and unstructured EHR features -- that are iteratively refined during training to improve model calibration. We evaluate WEST on two rare pulmonary diseases using EHR data from Boston Children's Hospital and show that it outperforms existing methods in phenotype classification, identification of clinically meaningful subphenotypes, and prediction of disease progression. By reducing reliance on manual annotation, WEST enables data-efficient rare disease phenotyping that improves cohort definition, supports earlier and more accurate diagnosis, and accelerates data-driven discovery for the rare disease community.

\vspace{1cm}

\textbf{Keywords}: Weakly Supervised Learning, Transformer Models, Rare Disease, Computational Phenotyping, Subphenotyping, Electronic Health Records

\newpage

\section{Introduction}  

Rare diseases are broadly defined as conditions affecting fewer than 1 in 2,000 people in any World Health Organization region or, in the United States, as those affecting fewer than 200,000 people \cite{health2024landscape, wang2024operational}. While individually uncommon, rare diseases collectively impose a substantial public health burden, affecting an estimated 300-400 million people worldwide \cite{health2024landscape, marwaha2022guide}. In the United States alone, approximately 30 million individuals -- 10\% of the population -- are living with a rare disease, a prevalence comparable to that of type 2 diabetes \cite{marwaha2022guide, boulanger2020establishing}.

Despite this widespread impact, rare diseases -- spanning more than 7,000 distinct conditions -- remain disproportionately difficult to diagnose. Many clinicians encounter some of these conditions only once, if ever, in their careers, limiting familiarity with the full spectrum of clinical presentations \cite{mak2024computer, rubinstein2020case}. These challenges contribute to the so-called ``diagnostic odyssey'' -- a years-long process marked by inconclusive tests, repeated specialist referrals, and frequent misdiagnoses -- that most rare disease patients endure \cite{bauskis2022diagnostic}. On average, patients consult between three and ten physicians and wait four to seven years before receiving a correct diagnosis \cite{stoller2018challenge, health2024landscape, mak2024computer}. Such delays prevent patients from receiving timely treatment, increase the risk of preventable complications, and contribute to premature mortality \cite{sreih2021diagnostic, gunne2020retrospective, mazzucato2023estimating}. The burden is particularly acute in pediatrics, as 70\% of rare diseases present in childhood and 30\% of affected children die before age five \cite{eclinicalmedicine2023rare, health2024landscape}. There is therefore an urgent need for earlier, more accurate diagnosis to improve outcomes and quality of life across the lifespan.

Diagnostic challenges are even more pronounced in rare pulmonary diseases, which are notoriously difficult to identify due to symptomatic overlap with common respiratory conditions. Up to one-third of individuals initially diagnosed with asthma are later found to have been misdiagnosed, with their symptoms attributable instead to less prevalent comorbid conditions \cite{gherasim2018confounders, kavanagh2019over}. Pulmonary hypertension (PH), a progressive disorder characterized by elevated pulmonary arterial pressure, frequently presents with nonspecific symptoms such as breathlessness, fatigue, and weakness -- features that closely resemble asthma \cite{ruopp2022diagnosis, galie20162015}. This overlap often delays recognition until irreversible vascular damage has occurred \cite{brown2011delay}. Severe asthma, a distinct and high-burden phenotype requiring high-dose inhaled corticosteroids plus a second controller, presents a similarly complex diagnostic challenge \cite{chung2018diagnosis}. Despite accounting for more than one-third of asthma-related deaths, severe asthma remains under-recognized, and its clinical heterogeneity further complicates timely diagnosis and effective management \cite{levy2014asthma}. Together, these pitfalls highlight the limitations of relying solely on clinical expertise and emphasize the need for data-driven approaches capable of detecting subtle, multi-dimensional patterns often missed in routine practice.

Efforts to consolidate rare disease cases into condition-specific registries have provided valuable research resources \cite{d2016creating}, but registries are often too small and narrow in scope to support large-scale clinical studies \cite{gliklich2014registries, hageman2023systematic}. Moreover, because inclusion requires a confirmed diagnosis, patients with atypical presentations or missed diagnoses -- those most crucial for building representative datasets -- are systematically excluded. The widespread adoption of electronic health records (EHRs) has enabled rare disease investigation at scale, capturing a broader spectrum of clinical presentations than traditional registries. EHRs contain rich longitudinal data in both structured (e.g., diagnosis, medication, and procedure codes) and unstructured (e.g., free-text notes from which concept unique identifiers [CUIs] can be extracted) formats, documenting a patient's diagnostic journey -- including misdiagnoses and testing patterns that can illuminate rare disease phenotypes \cite{garcelon2020electronic}. Leveraging these data, machine-assisted diagnostic approaches are increasingly integrated into research and clinical workflows, with notable success in pulmonary medicine: they support real-time imaging interpretation \cite{walsh2018deep, huang2019deep}, flag high-risk patients for specialist referrals \cite{kaplan2021artificial}, and retrospectively identify undiagnosed individuals for registries and observational studies \cite{geva2017computable}.

Central to such efforts is computational phenotyping, which seeks to automate the identification of disease patterns in EHR data and further stratify patients into clinically meaningful subgroups based on prognosis or treatment response. Traditionally, phenotyping has relied on rule-based algorithms that apply predefined logical criteria -- such as diagnostic codes, relevant medications, or abnormal lab values -- to infer disease status \cite{shivade2014review, alzoubi2019review}. While effective for well-characterized diseases with standardized coding, these methods perform poorly in rare diseases, which are clinically heterogeneous and often lack codified diagnostic criteria \cite{yang2023machine, banda2018advances}.

To address these limitations, machine learning (ML) and especially deep learning (DL) have emerged as powerful alternatives for large-scale phenotyping \cite{yang2023machine, callahan2023characterizing}. A key advance in this domain is representation learning, which transforms high-dimensional clinical data into lower-dimensional embeddings that preserve semantic and contextual relationships \cite{choi2016multi}. Within this framework, medical concepts from both structured and unstructured data are mapped to embeddings pre-trained on co-occurrence and semantic context \cite{weng2019representation}. These concept-level embeddings can then be aggregated into patient-level representations to support downstream phenotyping and subphenotyping tasks \cite{fridgeirsson2023attention}.

Despite their success in modeling common diseases, ML and DL methods often struggle in rare disease contexts due to both data and modeling challenges. EHR data are inherently high-dimensional, sparse, and noisy \cite{banerjee2023machine, schaefer2020use}. Importantly, the presence of a diagnostic code or concept in the EHR does not guarantee a confirmed diagnosis: codes may be entered for billing, used provisionally, or persist from outdated assessments \cite{yang2024noise, wu2020statistics}. For rare diseases, which are inconsistently documented and frequently underrecognized, these issues are particularly acute. Variability in documentation practices across providers and institutions further complicates the curation of accurate large-scale training sets. Together, these factors introduce substantial noise into downstream phenotyping. Compounding these challenges, most ML/DL methods for phenotyping rely on supervised learning, which requires large, high-quality labeled datasets -- a resource rarely available in rare disease research \cite{haneuse2020small}. Models trained on small gold-standard cohorts often overfit and fail to generalize, limiting clinical utility. These constraints have fueled growing interest in weakly supervised learning, which exploits large collections of noisy or partially labeled data to enhance model robustness under low-label conditions.

Among emerging DL architectures, transformers have shown particular promise for EHR phenotyping due to their ability to capture complex temporal dependencies and long-range relationships across irregular clinical events \cite{vaswani2017attention, yang2023transformehr}. Models such as BEHRT \cite{li2020behrt}, Med-BERT \cite{rasmy2021med}, RatchetEHR \cite{hirszowicz2024icu}, and Foresight \cite{kraljevic2024foresight} have achieved strong performance across predictive and classification tasks in both structured and unstructured data. These advances underscore the potential of transformer-based models for scalable phenotyping. Yet, despite their promise, transformer models remain constrained by the same supervised learning paradigm that limits other ML/DL applications in rare diseases. Existing approaches still require large volumes of clean, labeled data, limiting applicability in low-label, high-noise scenarios. Their potential for rare disease detection is therefore far from fully realized.

To bridge this gap, we propose a weakly supervised transformer (WEST) framework for rare disease phenotyping and subphenotyping from EHRs. The method combines a small set of expert-validated gold-standard labels with a much larger set of iteratively refined silver-standard labels derived from real-world EHR data. By integrating precise supervision with abundant but noisy signals, WEST learns robust patient representations optimized for downstream classification and clustering tasks. Importantly, the framework performs effectively even when only positive gold-standard cases are available, making it applicable in settings where registries exist but large-scale chart review is infeasible. Using PH and severe asthma as motivating case studies, we demonstrate that this approach improves phenotype detection and enables clinically meaningful subgroup discovery in real-world, data-limited settings.

\section{Methods}

\begin{figure}[H]
    \centering
    \includegraphics[width=0.9\linewidth]{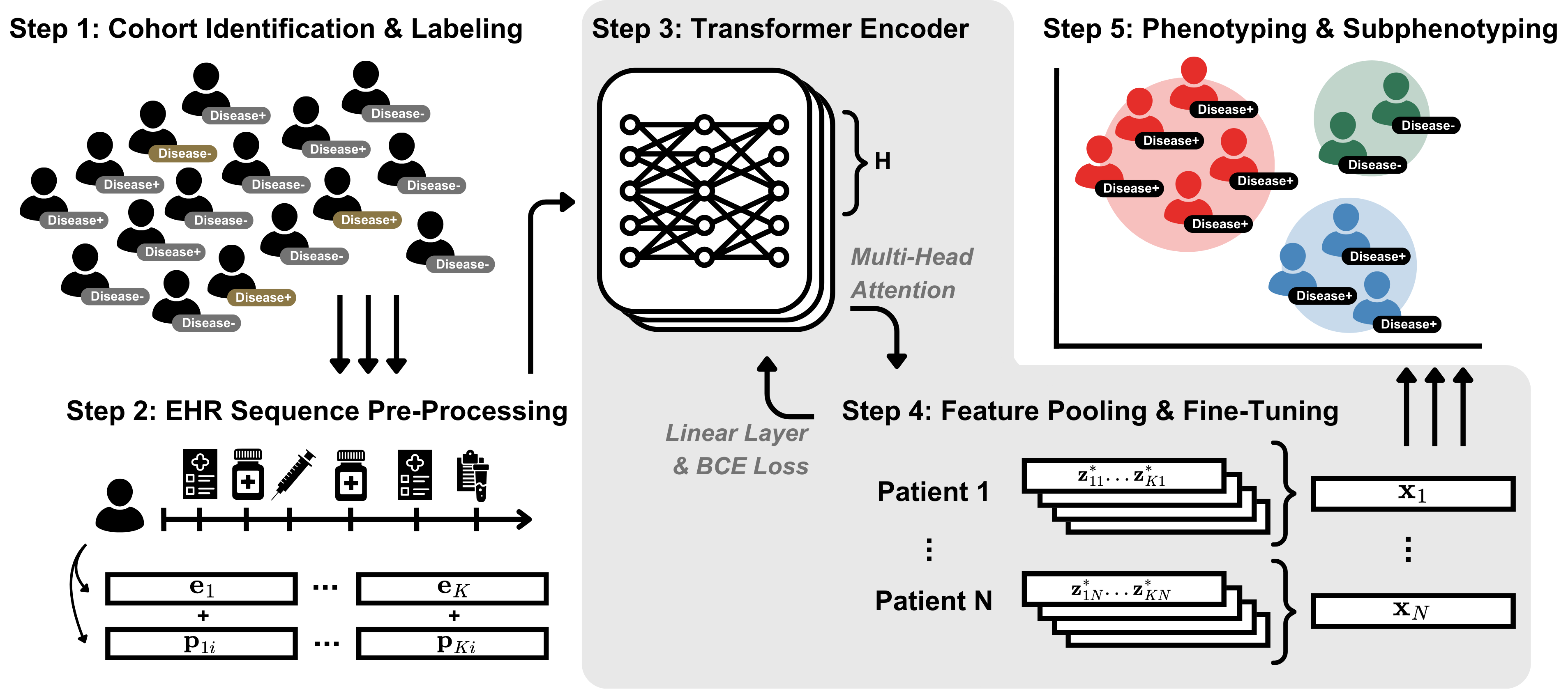}
    \caption{Overview of the WEST phenotyping pipeline.}
    \label{fig:overview}
\end{figure}

Our end-to-end WEST phenotyping pipeline integrates representation learning with weak supervision and iterative label refinement to enable data-efficient EHR phenotyping. We first identify a high-risk patient cohort and assign initial phenotypic labels using gold- or silver-standard sources (\S2.1). Each patient's longitudinal clinical history is then transformed into a structured input sequence through a multi-step pre-processing pipeline that includes event aggregation, feature selection, and frequency encoding (\S2.2). These inputs are processed by a multi-layer transformer encoder that models dependencies among clinical concepts (\S2.3). We then aggregate concept-level embeddings to generate patient-level representations, apply a classification head, and iteratively refine the silver-standard labels through weak supervision (\S2.4). The framework outputs both a patient-level phenotype prediction and a low-dimensional embedding suitable for clustering and visualization. An overview of the pipeline is shown in Figure~\ref{fig:overview}.

\subsection{Cohort Identification and Labeling}

We start by constructing a high-risk patient cohort -- individuals whose EHRs exhibit clinical features suggestive of the target condition or related conditions associated with elevated risk. For each disease-specific task, we designate a target diagnostic code or concept $c^*$, which serves as an anchor for identifying relevant features and guiding the label refinement process.

Let \( i = 1, \ldots, N \) index all patients in the high-risk cohort. Each patient \( i \) is assigned a label \( y_i \) reflecting their phenotype status. Based on the source and reliability of the label, patients are stratified into two cohorts:

\begin{enumerate}
    \item \textit{Gold-Standard Cohort:} Patients whose disease status has been confirmed through expert physician chart review or inclusion in a dedicated disease registry. These patients are assigned gold-standard labels, denoted \( y_i^{\text{gold}} \), which serve as high-fidelity references for model training and evaluation. We allow this set to be small to ensure that the WEST pipeline is label efficient. 
    
    \item \textit{Silver-Standard Cohort:} Patients with possible but unconfirmed diagnoses. These patients are assigned silver-standard labels, denoted \( y_i^{\text{silver}} \), inferred from the EHR data. Silver-standard labels can be defined using rule-based heuristics -- such as exceeding a threshold number of occurrences of \( c^* \) -- or derived from the probabilistic predictions of unsupervised automated phenotyping algorithms such as PheNorm \cite{phenorm} or KOMAP \cite{komap}. While these criteria expand the size of the labeled dataset, silver-standard labels are inherently noisier and require iterative refinement.
\end{enumerate}

The full set of training labels \( \{ y_i \} \) is drawn from both cohorts and defined as:

\begin{equation*}
    y_i = 
    \begin{cases}
      y_i^{\text{gold}}, & \text{if patient } i \text{ is in the gold-standard cohort}, \\
      y_i^{\text{silver}}, & \text{if patient } i \text{ is in the silver-standard cohort}.
    \end{cases}
\end{equation*}

A central component of our framework is the iterative refinement of silver-standard labels. Unlike gold-standard labels, which remain fixed, silver-standard labels are dynamically updated during model training. After each training round, the model generates updated predictions for the silver-standard cohort, and these predicted probabilities replace the previous labels. This weakly supervised approach allows the model to progressively improve label quality, enabling more accurate phenotype classification while leveraging the scale and diversity of real-world EHR data.

\subsection{EHR Sequence Pre-Processing}

We transform each patient's raw EHR into a structured representation suitable for transformer-based learning. This pre-processing pipeline comprises three key stages: (1) sequential representation of clinical histories, (2) label-aware augmentation for gold-standard patients, and (3) construction of input embeddings via feature selection and frequency encoding.

\subsubsection{Sequential Representation of EHR Data}

For each patient \( i \), the EHR is modeled as a temporal sequence of clinical events partitioned into discrete time windows. These windows reflect clinically meaningful periods such as visits, months, or hospitalization episodes. Let the patient sequence be:

\begin{equation*}
    \mathcal{P} = \{\mathcal{V}_1, \mathcal{V}_2, \dots, \mathcal{V}_T\},
\end{equation*}

where \( T \) is the number of observed time windows. Each window \( \mathcal{V}_t \) contains a set of documented medical concepts and their associated occurrence counts:

\begin{equation*}
    \mathcal{V}_t = \{(c_{t1}, n_{t1}), (c_{t2}, n_{t2}), \dots, (c_{tK_t}, n_{tK_t})\},
\end{equation*}

where \( c_{tk} \) denotes a medical concept and \( n_{tk} \) the number of times it was recorded in window \( \mathcal{V}_t \). The number of concepts \( K_t \) may vary across windows and patients.

\subsubsection{Label-Aware Augmentation for Gold-Standard Patients}

To enhance generalization and enable effective learning from high-quality labeled examples, we apply two augmentation strategies to the gold-standard cohort: oversampling and dynamic temporal truncation. These methods address class imbalance between gold- and silver-standard cohorts and introduce variability into training.

First, we mitigate the limited size of the gold-standard cohort by oversampling. Each gold-standard patient is replicated \( r \) times in the training data, ensuring that high-confidence examples are adequately represented and not overshadowed by the larger, noisier silver cohort. This increases the frequency with which the model encounters trusted labels during training, reinforcing supervision from reliable examples.

Second, we apply temporal truncation to simulate the incompleteness and variability typical of real-world EHRs. During each training iteration, for a patient sequence \( \mathcal{P} = \{\mathcal{V}_1, \dots, \mathcal{V}_T\} \), we randomly sample a start and end index, \( t_{\text{start}} \) and \( t_{\text{end}} \), such that \( 1 \leq t_{\text{start}} \leq t_{\text{end}} \leq T \). The truncated sequence is defined as:

\begin{equation*}
\mathcal{P}' = \{\mathcal{V}_{t_{\text{start}}}, \dots, \mathcal{V}_{t_{\text{end}}}\}.
\end{equation*}

This exposes the model to a variety of partial clinical trajectories -- some early, some late -- mimicking patients presenting at different disease stages or lacking complete documentation. Over time, this dynamic sampling increases the diversity of training examples derived from a fixed gold-standard set and improves robustness to temporal variability in real-world EHR data.

\subsubsection{Feature Engineering and Embedding Construction}

To prepare each sequence \( \mathcal{P} \) or its truncated version \( \mathcal{P}' \) as input to the transformer, we construct a structured representation through several pre-processing steps.

\subsubsection*{\textit{(a) Concept Aggregation and Pre-Trained Embeddings}}

Let \( \mathcal{C} = \{(c_1,n_1), (c_2,n_2), ..., (c_{K},n_{K}) \} \) denote the set of unique concepts and their cumulative counts across a patient's selected time period, whether from \( \mathcal{P} \) or \( \mathcal{P}' \). Each concept \( c_k \in \mathcal{C} \) is mapped to a vector representation \( \mathbf{e}_k \) using a pre-trained embedding model (PEM) for clinical concepts such as SapBERT \cite{liu2020sapbert}, CODER \cite{yuan2022coder}, MUGS \cite{li2024multi}, or ONCE \cite{komap}:

\begin{equation}
    \mathbf{e}_k = \text{PEM}(c_k), \quad \mathbf{e}_k \in \mathbb{R}^{d_{\text{input}}}.
\end{equation}

Since the transformer model operates in a hidden space of dimension \( d_{\text{model}} \), we project each embedding into this space via a learnable linear transformation:

\begin{equation}
    \mathbf{e}_k^{\text{proj}} = \mathbf{W}^{\text{proj}} \mathbf{e}_k + \mathbf{b}^{\text{proj}}, \quad \mathbf{e}_k^{\text{proj}} \in \mathbb{R}^{d_\text{model}},
\end{equation}

where \( \mathbf{W}^{\text{proj}} \in \mathbb{R}^{d_{\text{model}} \times d_{\text{input}}} \) and \( \mathbf{b}^{\text{proj}} \in \mathbb{R}^{d_{\text{model}}} \) are learnable parameters.

\subsubsection*{\textit{(b) Similarity-Based Feature Selection}}

Given the potentially large number of unique concepts in $\mathcal{C}$, we perform feature selection to retain only those most relevant to the target condition. This serves two purposes: (1) reducing noise from unrelated concepts, and (2) lowering computational burden, since transformer attention scales quadratically with the number of input tokens. To identify relevant features, we compute the cosine similarity between the embedding of each concept and that of the target concept $c^*$, representing the disease condition of interest:

\begin{equation}
    S(c_k, c^*) = \frac{\mathbf{e}_k \cdot \mathbf{e}^*}{\|\mathbf{e}_k\| \|\mathbf{e}^*\|},
\end{equation}

where \( \mathbf{e}_k \) and \( \mathbf{e}^* \) are the respective embeddings. The top \( K^* \) concepts with the highest similarity scores are retained:

\begin{equation*}
    \mathcal{C}^* = \{(c_1,n_1), (c_2,n_2), ..., (c_{K^*},n_{K^*}) \}, \quad \text{where } S(c_1, c^*) \geq S(c_2, c^*) \geq ... \geq S(c_{K^*}, c^*).
\end{equation*}

The target \( c^* \) is always included to ensure phenotype-specific information is preserved. Each \( n_k \) denotes the total count of concept \( c_k \) across all relevant time windows.

\subsubsection*{\textit{(c) Concept Frequency Encoding}}

At this stage, we have constructed an aggregated set $\mathcal{C}^*$ comprising unique clinical concepts and their corresponding cumulative frequencies, which summarize a patient's longitudinal medical history. To encode concept frequency -- serving as a proxy for clinical significance, capturing aspects such as chronicity or ongoing management -- we introduce a frequency-based embedding mechanism.

Each patient-specific cumulative count $n_{ki}$ for concept $c_k$ is projected into the model's embedding space through a two-layer feedforward network with a SwiGLU activation function, a gated variant of the linear unit shown to improve expressivity and training stability in transformer feedforward layers \cite{shazeer2020glu}:

\begin{equation}
    \mathbf{p}_{ki} = \mathbf{W}^{\text{pos}}_2
    \text{SwiGLU} \left( n_{ki} \mathbf{W}^{\text{pos}}_1 + \mathbf{b}^{\text{pos}}_1 \right)  + \mathbf{b}^{\text{pos}}_2, \quad \mathbf{p}_{ki} \in \mathbb{R}^{d_{\text{model}}},
\end{equation}

with learnable parameters:

\[
\mathbf{W}^{\text{pos}}_1 \in \mathbb{R}^{\frac{d_{\text{model}}}{2} \times 1}, \quad
\mathbf{W}^{\text{pos}}_2 \in \mathbb{R}^{d_{\text{model}} \times \frac{d_{\text{model}}}{2}}, \quad
\mathbf{b}^{\text{pos}}_1 \in \mathbb{R}^{\frac{d_{\text{model}}}{2}}, \quad
\mathbf{b}^{\text{pos}}_2 \in \mathbb{R}^{d_{\text{model}}}.
\]

Unlike traditional positional encodings used in natural language processing (NLP), this representation is grounded in concept frequency rather than token order, offering a tailored signal for clinical models sensitive to the recurrence and persistence of medical events.

\subsubsection*{\textit{(d) Transformer Input Sequence}}

The final representation of each selected concept is obtained by summing its embedding and patient-specific frequency encoding:

\vspace{0.5em}

\begin{equation}
    \mathbf{z}_{ki} = \mathbf{e}_{k}^{\text{proj}} + \mathbf{p}_{ki}, \quad \mathbf{z}_{ki} \in \mathbb{R}^{d_{\text{model}}}.
\end{equation}

Here, \( \mathbf{z}_{ki} \) is the input token for concept \( c_k \) for patient $i$ to the transformer. If concept \( c_k \) is not observed for patient $i$, we set \( \mathbf{z}_{ki} =0\). This formulation allows the model to simultaneously capture semantic similarity across medical concepts and their implicit clinical significance based on frequency. The final patient sequence is:

\begin{equation*}
\mathbf{Z}_i = \{\mathbf{z}_{1i}, \mathbf{z}_{2i}, ..., \mathbf{z}_{K^*i}\}.
\end{equation*}

\subsection{Transformer Encoder}

Our model builds on a multi-layer transformer encoder but adapts it for the challenges of weakly supervised phenotyping. The encoder serves two purposes simultaneously: (1) patient-level classification, where the model predicts the probability that a patient has the target condition, and (2) representation learning, where it generates low-dimensional embeddings useful for clustering, subphenotyping, and visualization.

Each patient sequence $\mathbf{Z}_i$ is processed through stacked transformer encoder layers. Within each layer, multi-head self-attention models dependencies among medical concepts, enabling the network to focus on the parts of the record most informative for the target disease. Standard architectural elements -- including residual connections, layer normalization, and feedforward networks with nonlinear activations -- are incorporated to ensure stable training. Full mathematical details are provided in Section \textbf{S1} of the supplementary materials, which describe the multi-layer transformer architecture employed by WEST. The derivations clarify the inner workings of the transformer encoder, including its attention mechanism, projection layers, and feedforward components.


\subsection{Feature Pooling and Fine Tuning}

After passing through multiple transformer layers, the sequence of contextualized embeddings is aggregated into a fixed-length patient representation using mean pooling:

\begin{equation}
    \mathbf{x}_{i} = \frac{1}{K^*} \sum_{k=1}^{K^*} \mathbf{z}^*_{k}.
\end{equation}

This approach allows the model to capture contributions from all medical concepts while accommodating sequences of varying lengths. The pooled patient representation \( \mathbf{x}_i \) is passed through a classification head -- a linear layer followed by a sigmoid activation -- to produce a probability score:

\begin{equation}
    p(y_i) = \sigma(\mathbf{W}^{\text{class}} \mathbf{x}_i + \mathbf{b}^{\text{class}}),
\end{equation}

where \( \mathbf{W}^\text{class} \) and \( \mathbf{b}^\text{class} \) are learnable parameters. The sigmoid function \( \sigma(\cdot) \) maps the logit to a probability in the range \((0, 1)\). Model training employs binary cross-entropy (BCE) loss for classification. After each training round, the best-performing model on the validation set is used to update silver-standard labels using its predicted probabilities:

\begin{equation}
    y_i^{\text{silver}} \leftarrow  p(y_i).
\end{equation}

This iterative label refinement allows the model to incorporate its own predictions, progressively improving phenotype classification over training cycles.

\subsection{Real World Validation}
We evaluated our WEST framework on phenotyping two rare pulmonary diseases, PH and severe asthma, using EHR data from Boston Children's Hospital. For each disease, the model was trained and validated independently using disease-specific EHR cohorts and labels curated by domain experts via chart review.

\subsubsection{Data Curation} 

For both PH and severe asthma, we constructed disease-specific cohorts by first identifying at-risk patient populations from the EHR data at Boston Children's Hospital. The at-risk PH cohort comprised 14,305 randomly selected patients with PheCode 415.2 (indicative of potential PH), while the severe asthma cohort comprised 7,822 randomly selected patients with PheCodes beginning with J45 (indicative of asthma of any severity).

Gold-standard cohorts consisted of patients with a confirmed diagnosis, established either through expert chart review or enrollment in a disease-specific registry. The PH gold-standard cohort comprised 531 patients, with 106 (20\%) set aside for validation and testing, while the severe asthma cohort comprised 248 patients, with 99 (40\%) set aside. These held-out patients were further split into two equally sized cross-validation folds: one used for validation (model checkpoint selection) and the other for testing (performance evaluation). Final performance metrics were averaged across cross-validation folds.

The silver-standard cohorts comprised the remaining at-risk patients whose phenotype status had not been definitively adjudicated -- 13,774 for PH and 7,575 for severe asthma. Initial probabilistic labels $y_i^{\text{silver}} \in (0, 1)$ were assigned to these patients using the KOMAP algorithm \cite{komap}.

For PH, KOMAP was applied to codified EHR features, including PheCode diagnoses, RxNorm medications, and Clinical Classifications Software (CCS) procedure codes. For severe asthma, KOMAP was applied to natural language features extracted from clinical notes using Narrative Information Linear Extraction (NILE) \cite{yu2013nile}. From these codes and concepts, we designated PheCode:415.2 for PH and CUI:C0581126 for severe asthma as the target phenotypes. For representation learning, we mapped the codified EHR data in the PH cohort to pre-trained MUGS embeddings \cite{li2024multi} and the NLP-derived features in the severe asthma cohort to pre-trained ONCE embeddings \cite{komap}.

\subsubsection{Validation Metrics}

We first assessed the classification performance of the WEST pipeline on labels not used during training. Evaluation was performed with two-fold cross-validation, computing AUC, positive predictive value (PPV), and specificity for each fold and then averaging across folds. Sensitivity was fixed at 80\%. Performance was compared against five baselines: 

\begin{enumerate}
    \item \textbf{Count}: Frequency of the target concept appearing in each patient's EHR.
    \item \textbf{KOMAP}: The initial silver-standard probabilities generated by KOMAP \cite{komap}.
    \item \textbf{XGBoost}: A supervised gradient-boosted trees classifier \cite{chen2016xgboost}.
    \item \textbf{Transformer (silver = gold)}: A transformer trained by treating all silver-standard labels as gold-standard, without any iterative updates or data augmentation.
    \item \textbf{Transformer (gold only)}: A transformer trained solely on gold-standard labels.
\end{enumerate}

As additional ablation studies, we examined two aspects of gold-standard supervision. First, we varied the number of gold-standard labels used for training, gradually increasing the labeled set from 25 to 400 examples. Second, we modified WEST to train without any gold-standard negative labels, simulating a setting where no confirmed negatives are available and all negative training samples are drawn from the silver-standard cohort.

We next evaluated whether the learned patient representations captured clinically meaningful heterogeneity. Using patients with known disease status who were not included in training, we tested whether the embeddings could distinguish true positive from true negative cases. To visualize this separation, we applied t-distributed stochastic neighbor embedding (t-SNE) to the WEST embeddings of held-out patients and compared the resulting visualization with that obtained from embeddings generated using term frequency-inverse document frequency (TF-IDF), a commonly used feature engineering approach \cite{ramos2003using, van2008visualizing}.

For subphenotype discovery, we focused on patients predicted to have a positive disease status. We first reduced the learned embeddings using principal component analysis (PCA), retaining components that together explained at least 90\% of the variance. We then applied k-means clustering to identify clinically meaningful patient subgroups. Finally, we evaluated the prognostic relevance of these clusters: in PH, by comparing survival distributions using Kaplan-Meier curves; and in severe asthma, by estimating hazard ratios (HRs) for signs and symptoms indicative of disease severity across clusters.

\subsection{Hyperparameter Tuning}

We performed hyperparameter tuning using a 2-fold cross-validation procedure to robustly select model configurations. For each hyperparameter setting, the model was trained on one fold and evaluated on the other. A random search strategy was employed to explore the following hyperparameter space: batch size $\in \{64, 128, 256\}$, learning rate $\in \{5\text{e-}4, 1\text{e-}3, 2\text{e-}3\}$, hidden dimension $\in \{32, 64, 128\}$, number of Transformer layers $\in \{2, 3, 4\}$, dropout rate $\in \{0.3, 0.7\}$, and number of training epochs $\in \{30, 50\}$. The area under the receiver operating characteristic curve (AUC) served as the primary selection metric, and the chosen hyperparameters for each fold were subsequently used to train the final models.

\section{Results}

\subsection{Pulmonary Hypertension}

\subsubsection{Classification Performance}

As shown in Table~\ref{tab:komap_metrics1}, the WEST pipeline trained with both positive and negative gold-standard labels achieved the highest AUC (0.93), PPV (0.95), and specificity (0.92). WEST trained without gold-standard negative labels also outperformed all baselines.  

\begin{table}[H]
\centering
\begin{tabular}{|l|c|c|c|c|c|c|c|}
\hline
\textbf{Metric} & \textbf{Count} & \textbf{KOMAP} & \textbf{XGBoost} & \makecell{\textbf{Transformer} \\ \small{(silver = gold)}} & \makecell{\textbf{Transformer} \\ \small{(gold only)}} & \makecell{\textbf{WEST} \\ \small{(w/o neg)}} & \makecell{\textbf{WEST} \\ \small{(w/ neg)}} \\ 
\hline
AUC & 0.85 & 0.86 & 0.82 & 0.84 & 0.88 & 0.91 & \textbf{0.93} \\ 
\hline
PPV & 0.65 & 0.88 & 0.87 & 0.84 & 0.89 &  0.91 & \textbf{0.95} \\ 
\hline
Specificity & 0 & 0.78 & 0.76 & 0.70 & 0.81 & 0.84 & \textbf{0.92}  \\ 
\hline
\end{tabular}
\caption{Phenotype classification performance for PH. Transformer and WEST metrics are averaged across two cross-validation folds.}
\label{tab:komap_metrics1}
\end{table}

\begin{figure}[H]
    \centering
    \includegraphics[width=0.7\linewidth]{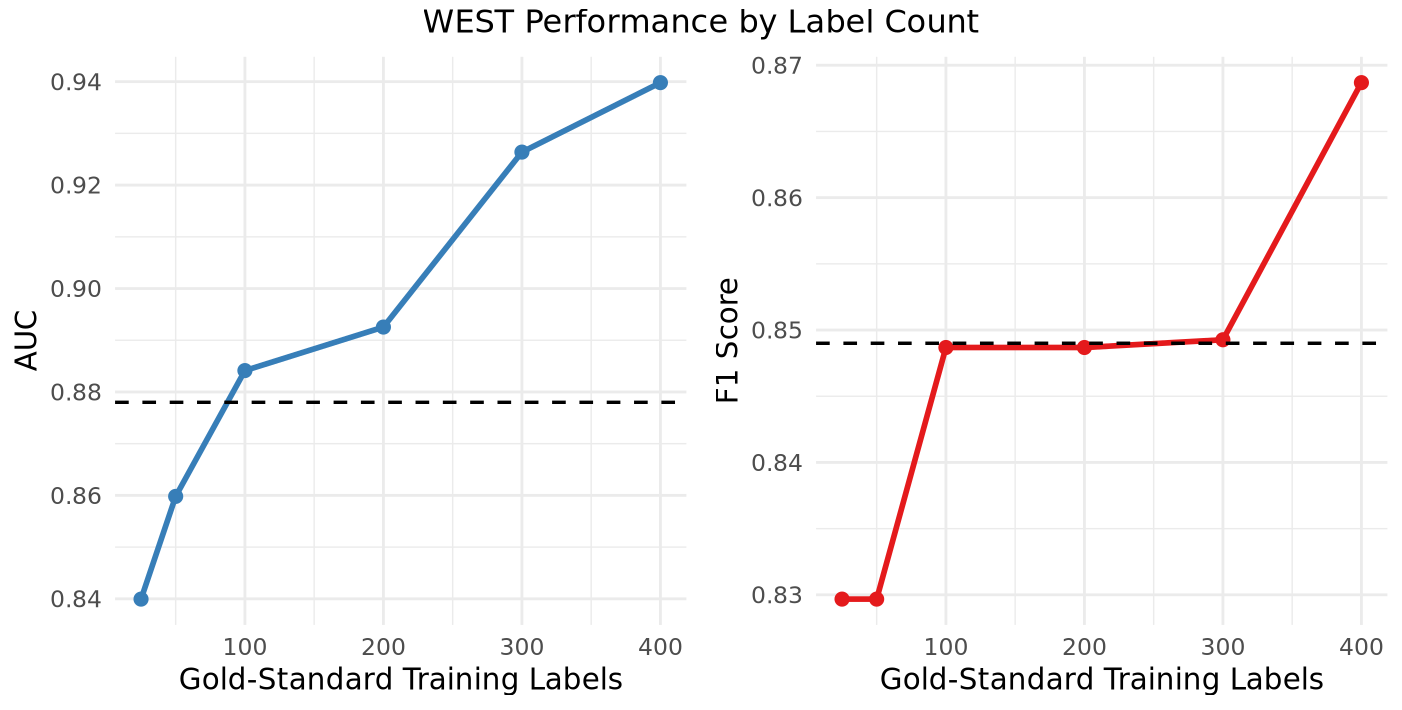}
    \caption{WEST performance as a function of the number of gold-standard training labels. Curves report AUC and F1 score. The horizontal dotted line represents the best performing baseline, \emph{Transformer (gold only)}.}
    \label{fig:ablation}
\end{figure}

\begin{figure}[H]
    \centering
    \includegraphics[width=0.7\linewidth]{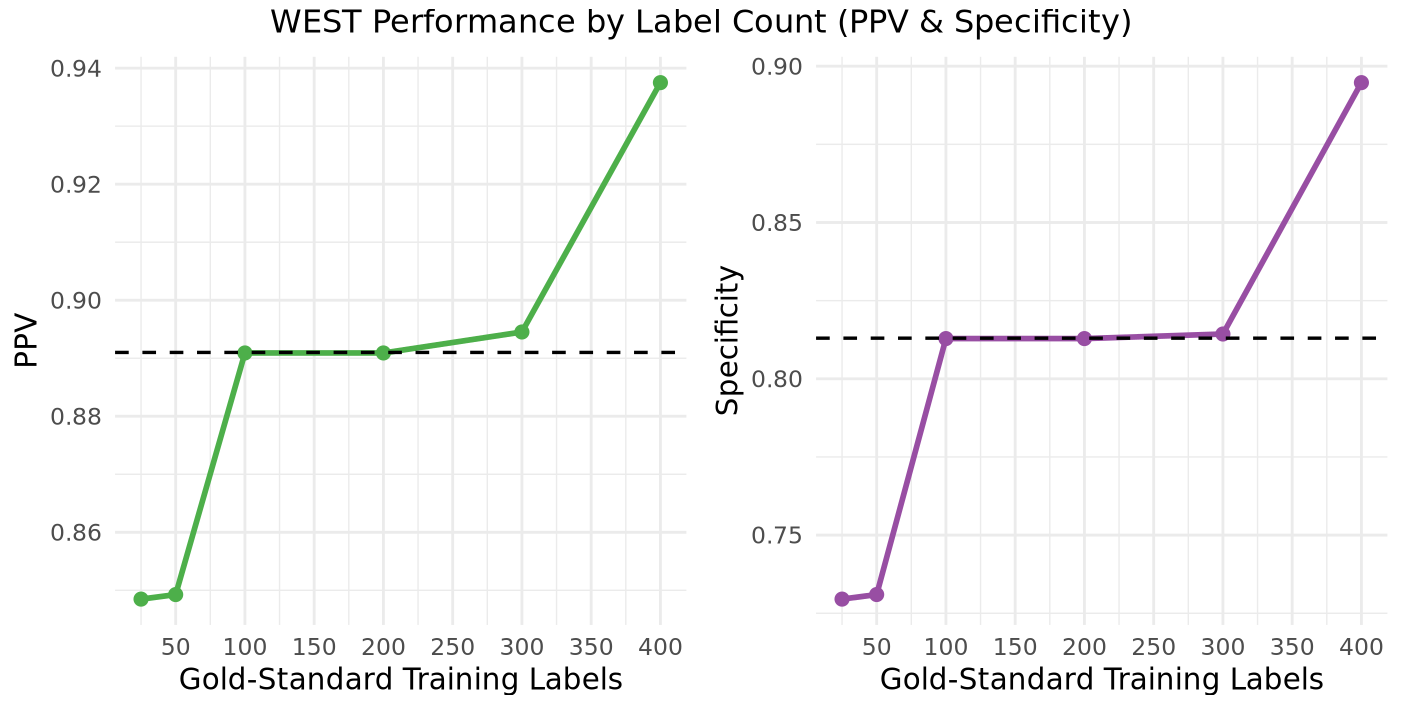}
    \caption{WEST performance as a function of the number of gold-standard training labels. Curves report PPV and specificity. The horizontal dotted line represents the best performing baseline, \emph{Transformer (gold only)}.}
    \label{fig:ablation_appendix}
\end{figure}

Figures~\ref{fig:ablation} and~\ref{fig:ablation_appendix} demonstrate that WEST performance increased with the number of gold-standard training labels. Notably, with only 100 labels, WEST already matched or surpassed all baseline methods, with further gains achieved at larger label sets.

\subsubsection{Clustering Performance}

As shown in Figure~\ref{fig:tsne}, PH-positive and PH-negative patients were more distinctly separated in the latent space using WEST embeddings than with TF-IDF.  

\begin{figure}[H]
    \centering
    \includegraphics[width=0.9\linewidth]{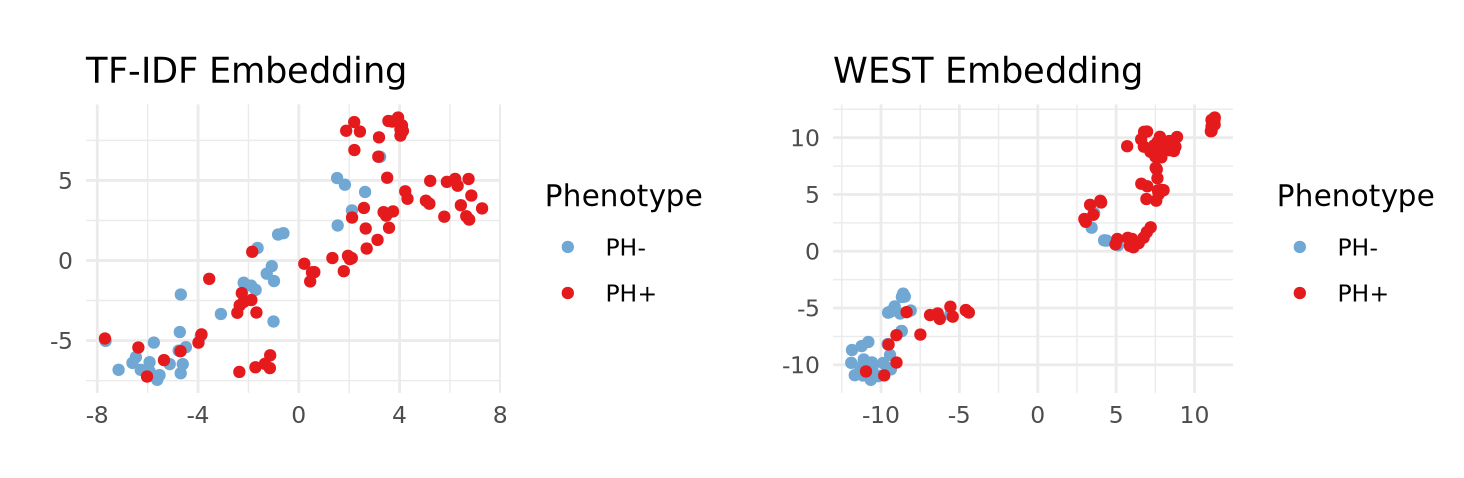}
    \caption{t-SNE visualization of patient-level embeddings for PH phenotypes using TF-IDF and WEST-derived representations.}
    \label{fig:tsne}
\end{figure}

\begin{figure}[H]
    \centering
    \includegraphics[width=0.7\linewidth]{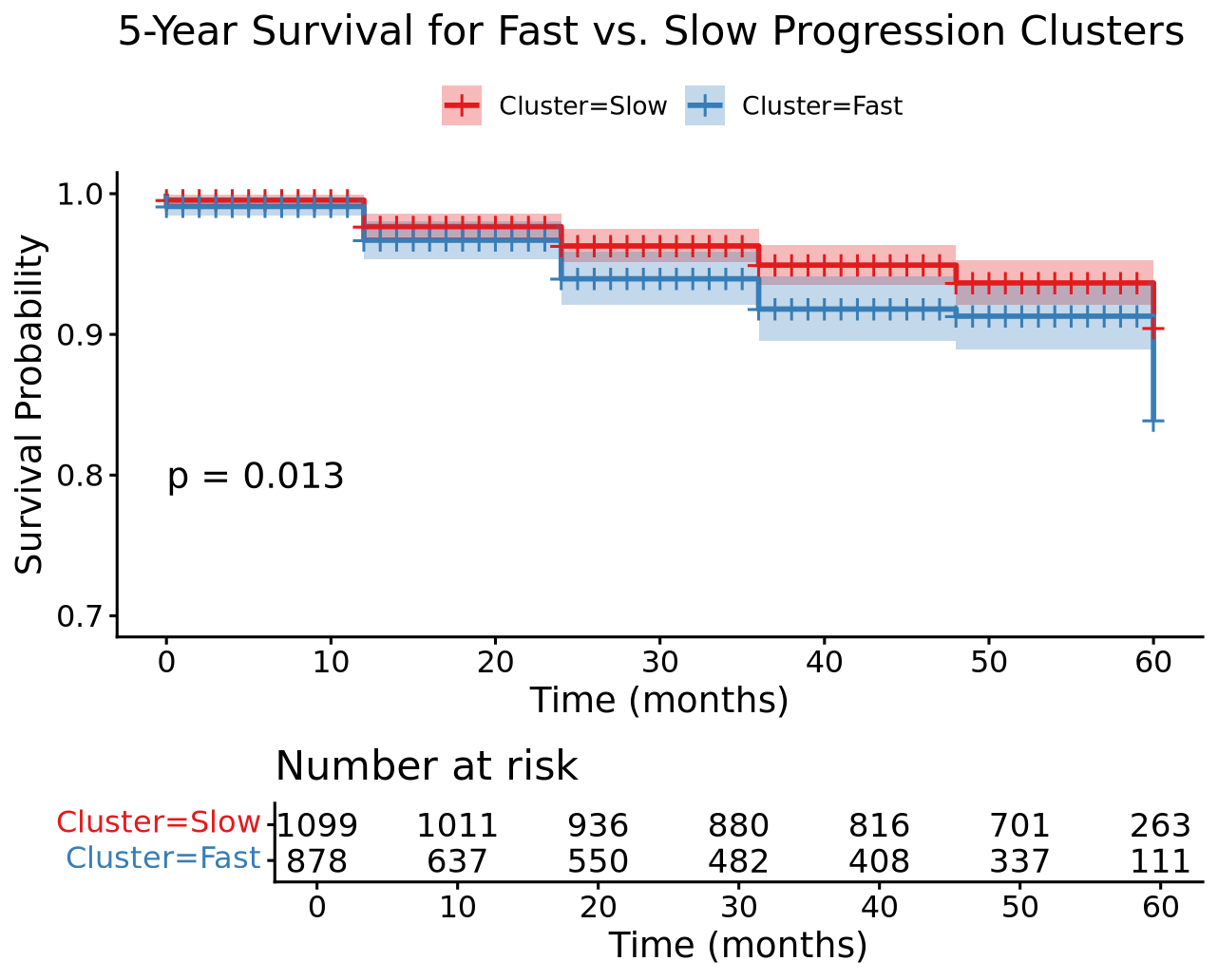}
    \caption{Kaplan-Meier survival curves for PH subgroups identified via k-means clustering of transformer embeddings.}
    \label{fig:survival}
\end{figure}

The WEST pipeline identified 1977 patients with PH. Clustering of PH-positive patient embeddings revealed two subgroups: a \textit{Slow Progression} cluster (\( n = 1099 \)) and a \textit{Fast Progression} cluster (\( n = 878 \)). Kaplan-Meier survival analysis showed a significant difference in 5-year mortality between the two clusters (\( p = 0.013 \); Figure~\ref{fig:survival}).

\subsection{Severe Asthma}
    
\subsubsection{Classification Performance}

Table~\ref{tab:komap_metrics2} presents classification metrics across methods. When trained with both positive and negative gold-standard labels, WEST achieved the highest AUC (0.87), PPV (0.80), and specificity (0.76) compared against all baselines. 

Patients predicted to have severe asthma showed increased risk for multiple indicators of disease severity compared with those predicted to have non-severe asthma. Significant associations were observed for recurrent status asthmaticus (HR = 55.30, 95\% CI: 43.93-69.61, $p < 0.0001$), respiratory failure (HR = 3.19, 95\% CI: 2.05-4.97, $p < 0.0001$), low oxygen events including hypoxia and desaturation (HR = 2.66, 95\% CI: 2.05-3.45, $p < 0.0001$), tachypnea (HR = 3.67, 95\% CI: 3.17-4.26, $p < 0.0001$), bronchospasm (HR = 3.49, 95\% CI: 2.20-5.55, $p < 0.0001$), and dyspnea (HR = 2.97, 95\% CI: 2.66-3.32, $p < 0.0001$) (Figure~\ref{fig:sa_hr_new}).

\begin{table}[H]
\centering
\begin{tabular}{|l|c|c|c|c|c|c|c|}
\hline
\textbf{Metric} & \textbf{Count} & \textbf{KOMAP} & \textbf{XGBoost} & \makecell{\textbf{Transformer} \\ \small{(silver = gold)}} & \makecell{\textbf{Transformer} \\ \small{(gold only)}} & \makecell{\textbf{WEST} \\ \small{(w/o neg)}} & \makecell{\textbf{WEST} \\ \small{(w/ neg)}} \\ 
\hline
AUC & 0.80 & 0.82 & 0.83 & 0.82 & 0.78 & 0.86 & \textbf{0.87} \\ 
\hline
PPV & 0.74 & 0.74 & 0.78 & 0.70 & 0.69 & 0.75 & \textbf{0.80} \\ 
\hline
Specificity & 0.68 & 0.66 & 0.72 & 0.60 & 0.60 & 0.69 & \textbf{0.76} \\ 
\hline
\end{tabular}
\caption{Phenotype classification performance for severe asthma. Transformer metrics are averaged across two cross-validation folds.}
\label{tab:komap_metrics2}
\end{table}

\subsubsection{Clustering Performance}

Again, patients with and without severe asthma were more distinctly separated in the latent space when using WEST embeddings compared to TF-IDF embeddings (Figure~\ref{fig:tsne}).  

\begin{figure}[H]
    \centering
    \includegraphics[width=0.9\linewidth]{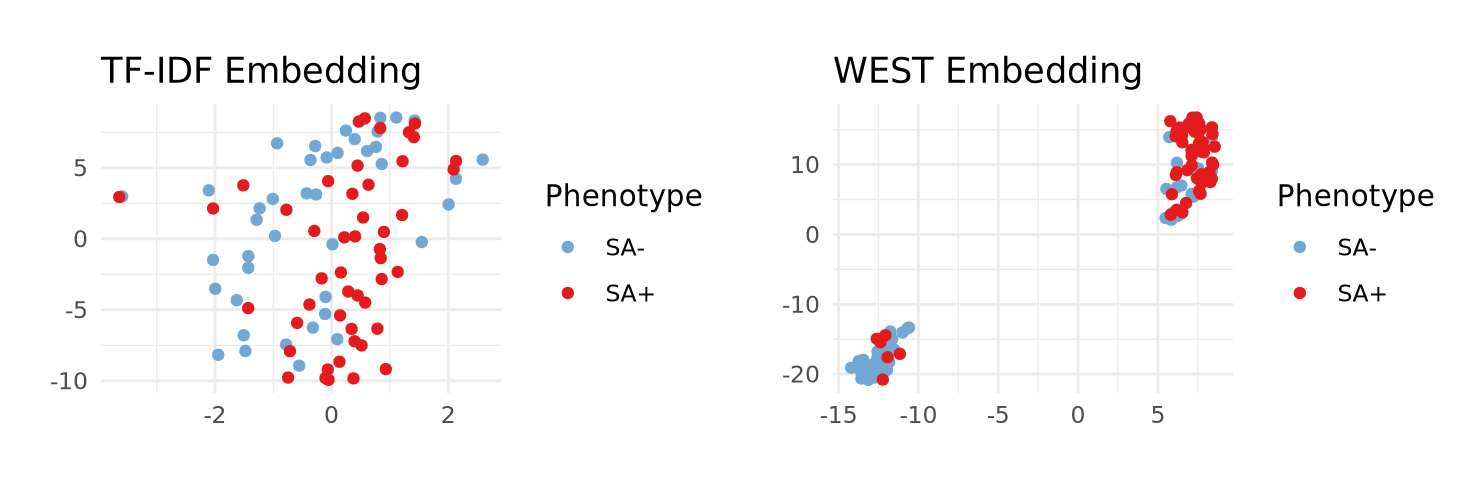}
    \caption{t-SNE visualization of patient-level embeddings for severe asthma (SA) phenotypes using TF-IDF and WEST-derived representations.}
    \label{fig:tsne}
\end{figure}

\begin{figure}[H]
    \centering
    \includegraphics[width=0.9\linewidth]{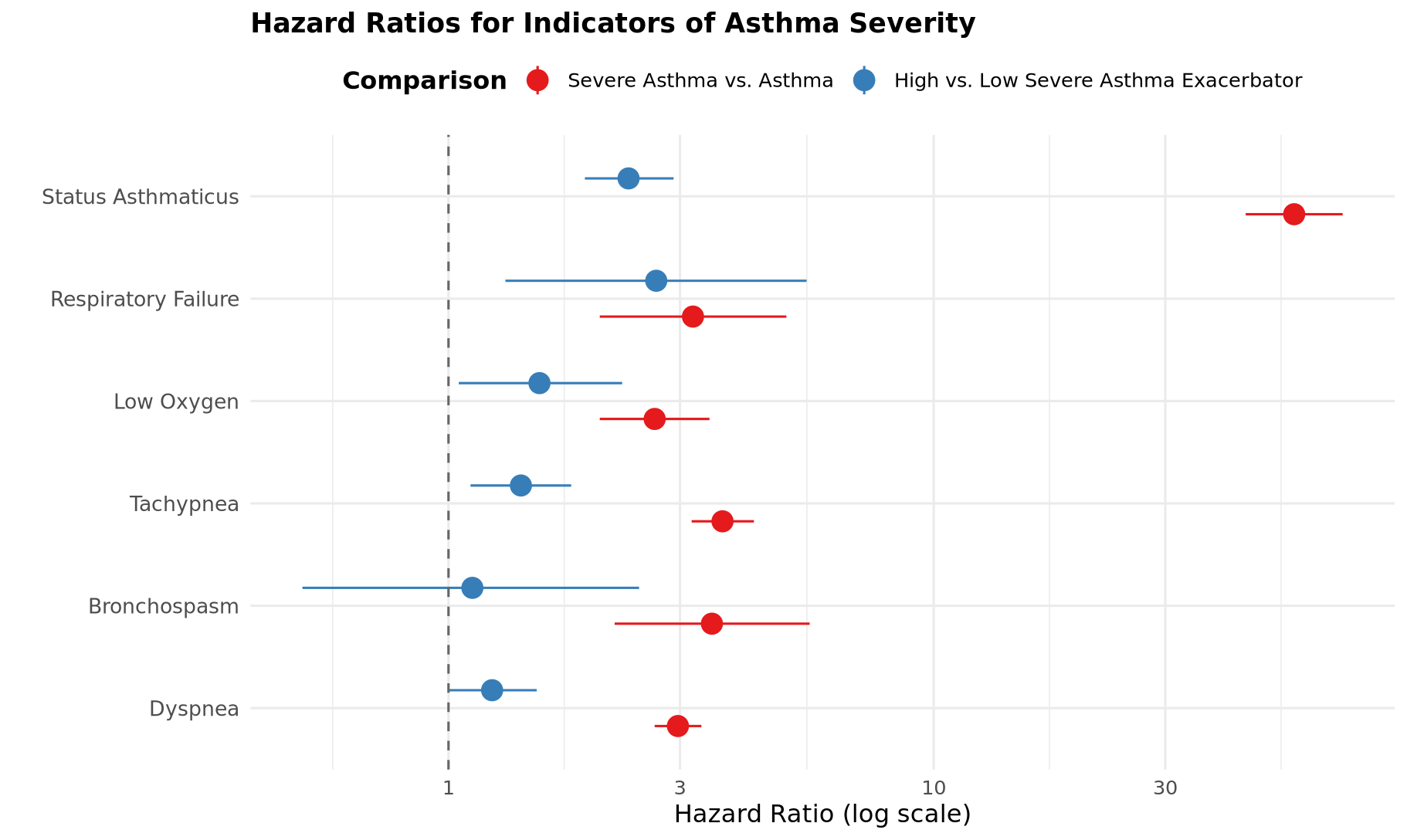}
    \caption{Hazard ratios and 95\% CIs for adverse events associated with asthma severity and exacerbation subphenotypes.}
    \label{fig:sa_hr_new}
\end{figure}

Among 582 patients predicted to have severe asthma, k-means clustering identified a \textit{Low Exacerbator} cluster ($n$ = 209) and a \textit{High Exacerbator} cluster ($n$ = 373). Patients in the High Exacerbator cluster had higher risk of recurrent status asthmaticus (HR = 2.35, 95\% CI: 1.91-2.91, $p < 0.0001$), respiratory failure (HR = 2.68, 95\% CI: 1.31-5.47, $p = 0.0068$), low oxygen events (HR = 1.54, 95\% CI: 1.05-2.28, $p = 0.0291$), and tachypnea (HR = 1.41, 95\% CI: 1.11-1.79, $p = 0.0050$) compared with the Low Exacerbator cluster (Figure~\ref{fig:sa_hr_new}).

\section{Discussion}

In this study, we introduce WEST, a weakly supervised transformer framework that integrates gold-standard annotations with iteratively refined silver-standard labels to enable data-efficient rare disease phenotyping from heterogeneous EHR data. Across PH and severe asthma case studies, WEST consistently outperformed rule-based and conventional machine learning baselines, producing patient representations that more clearly separated disease states and revealed clinically meaningful subgroups with divergent outcomes.

Among PH patients, clustering of WEST embeddings uncovered two distinct subgroups that we termed the \textit{Slow Progression} and \textit{Fast Progression} clusters. Patients in the latter group exhibited significantly poorer survival, highlighting clinically relevant heterogeneity in disease trajectories that may not be captured by current diagnostic labels. In severe asthma, clustering differentiated \textit{Low Exacerbator} and \textit{High Exacerbator} subgroups, the latter showing increased risks of recurrent status asthmaticus, respiratory failure, hypoxemia, and tachypnea. These patterns suggest that WEST-derived representations encode latent disease biology and care trajectories that extend beyond codified diagnoses. Such properties position WEST as a potential foundation for precision registry curation, proactive monitoring, and targeted intervention, enabling earlier identification of high-risk subgroups and more efficient allocation of clinical resources.

Methodologically, our work contributes to the growing literature on weak supervision and transformer-based modeling in healthcare, demonstrating that iterative silver-standard label refinement can effectively reduce noise and improve calibration during training. This label-efficient paradigm balances the breadth of routinely collected EHR data with the reliability of expert-validated annotations, yielding performance gains over both silver-only and fully supervised baselines. The superiority of transformer embeddings over TF-IDF representations further underscores the value of contextual modeling: by capturing temporal and semantic dependencies among codes, WEST generates clinically coherent patient spaces that facilitate downstream discovery, risk stratification, and hypothesis generation.

Several limitations merit discussion. First, our evaluation was retrospective and based on data from a single health system, which may limit generalizability to other care settings or populations. Second, while WEST integrates structured and unstructured data, the current implementation focuses on single-phenotype classification. Future extensions to multi-task learning and cross-institutional training could enhance scalability and robustness, particularly for ultra-rare diseases. Lastly, prospective validation and clinician-in-the-loop evaluation will be essential to assess real-world utility and interpretability in clinical workflows.

In summary, WEST advances weakly supervised learning for rare disease phenotyping by coupling transformer-based representation learning with iterative silver-label refinement. This framework improves data efficiency, enhances phenotype separability, and captures latent clinical heterogeneity, offering a scalable approach to harness EHR data for precision discovery and translational impact.

\section{Conclusion}

This study demonstrates that integrating weak supervision with transformer-based modeling enables scalable and data-efficient phenotyping for rare diseases. By coupling limited expert annotations with iteratively refined supervision, the framework achieves robust performance while capturing clinically meaningful heterogeneity within PH and severe asthma cohorts. These findings illustrate how weakly supervised transformers can complement traditional EHR-based phenotyping approaches by leveraging routine clinical data to reveal structure that extends beyond diagnostic codes and by supporting more consistent identification of patients with underrecognized or misclassified rare diseases.

More broadly, this work illustrates a general strategy for studying conditions that are too rare or heterogeneous for conventional supervised learning. Rather than replacing expert review, weakly supervised transformers can serve as augmentation tools -- helping prioritize cases for validation, enrich registries, and identify underrecognized subgroups for further study. Embedding such approaches within data curation and clinical research pipelines can accelerate discovery, enhance diagnostic accuracy, and improve the representativeness of real-world evidence for rare disease populations.

\section{Reproducibility}

\subsection{Data Availability}

The EHR data used in this study were obtained from Boston Children's Hospital and contain protected health information that cannot be shared publicly due to patient privacy regulations and institutional data use agreements. Access to these data is therefore restricted and cannot be distributed outside the institution. Derived, de-identified summary results supporting the findings of this study are available from the corresponding author upon reasonable request and subject to institutional approval.

\subsection{Code Availability}

Python implementation of the methodology developed and used in this study is available on GitHub at https://github.com/kfgreco/WEST.



\newpage

\printbibliography

\newpage

\textbf{\Large{Supplementary Materials}}

\setcounter{section}{0}
\renewcommand{\thesection}{S\arabic{section}}

\numberwithin{equation}{section}
\setcounter{equation}{0}

\section{Mathematical Details for Transformer Encoder}

Here, we provide additional mathematical details on the multi-layer transformer architecture employed by WEST. These derivations clarify the inner workings of the transformer encoder, including its attention mechanism, projection layers, and feedforward components.

\subsection{Multi-Head Attention}

Each transformer encoder layer applies multi-head self-attention to capture interactions between medical concepts. Given input embeddings \( \mathbf{z}_{ki} \in \mathbf{Z}_i \), the model computes queries, keys, and values using learnable projection matrices \( \mathbf{W}_Q, \mathbf{W}_K, \mathbf{W}_V \in \mathbb{R}^{d_{\text{model}} \times d_{\text{model}}} \). The self-attention mechanism is split into \( H \) heads, each operating in a subspace of dimension \( d_h = \frac{d_{\text{model}}}{H} \). For each head \( h \in \{1, ..., H\} \), input embeddings are projected as:

\begin{equation}
    \mathbf{q}_{ki}^{(h)} = \mathbf{W}_Q^{(h)} \mathbf{z}_{ki}, \quad \mathbf{k}_{ki}^{(h)} = \mathbf{W}_K^{(h)} \mathbf{z}_{ki}, \quad \mathbf{v}_{k}^{(h)} = \mathbf{W}_V^{(h)} \mathbf{e}_{k}^{\text{proj}},
\end{equation}

This multi-head design enables the model to attend to diverse contextual patterns across the input sequence. Attention weights within each head are computed using scaled dot-product attention:

\begin{equation}
    \text{score}_{k j i}^{(h)} = \frac{\mathbf{q}_{ki}^{(h)} \cdot \mathbf{k}_{ji}^{(h)}}{\sqrt{d_h}},
\end{equation}

followed by normalization with the softmax function:

\begin{equation}
    \alpha_{k j i}^{(h)} = \text{softmax}(\text{score}_{k j i}^{(h)}),
\end{equation}

which determines the influence of concept \( j \) on concept \( k \). The attention-based output for each head is:

\begin{equation}
    \mathbf{z}_{ki}^{(h)} = \sum_{j} \alpha_{k ji}^{(h)} \mathbf{v}_{j}^{(h)}.
\end{equation}

\subsection{Concatenation and Linear Projection}

Outputs from all attention heads are concatenated to restore the full embedding dimension:

\begin{equation}
    \mathbf{z}_{ki}^{\text{attn}} = \text{Concat}(\mathbf{z}_{ki}^{(1)}, \dots, \mathbf{z}_{ki}^{(H)}), \quad \mathbf{z}_{ki}^{\text{attn}} \in \mathbb{R}^{d_{\text{model}}}.
\end{equation}

A final linear projection aggregates information across heads:

\begin{equation}
    \mathbf{z}_{ki}^{\text{out}} = \mathbf{W}^{\text{out}} \mathbf{z}_{ki}^{\text{attn}},
\end{equation}

where \( \mathbf{W}^{\text{out}} \in \mathbb{R}^{d_{\text{model}} \times d_{\text{model}}} \) is a learnable weight matrix. To promote stable training and gradient flow, we apply a residual connection followed by layer normalization:

\begin{equation}
    \mathbf{z}_{ki}^\text{norm} = \text{LayerNorm}(\mathbf{z}_{ki} + \text{Dropout}(\mathbf{z}_{ki}^{\text{out}})).
\end{equation}

\subsection{Feedforward Network with SwiGLU Activation}

Each transformer layer includes a position-wise feedforward network with two linear layers and a SwiGLU activation:

\begin{equation}
    \mathbf{z}_{ki}^{\text{ffn}} = \mathbf{W}^{\text{ffn}}_2 \text{SwiGLU}( \mathbf{W}^{\text{ffn}}_1 \mathbf{z}_{ki}^{\text{norm}} + \mathbf{b}^{\text{ffn}}_1 ) + \mathbf{b}^{\text{ffn}}_2.
\end{equation}

A second residual connection and layer normalization step complete the transformer block:

\begin{equation}
    \mathbf{z}^*_{ki} = \text{LayerNorm}(\mathbf{z}^\text{norm}_{ki} + \text{Dropout}(\mathbf{z}_{ki}^{\text{ffn}})).
\end{equation}

\section{Supplementary Ablation Results (PPV and Specificity)}

\end{document}